# A Review of Predictive and Contrastive Self-supervised Learning for Medical Images

Wei-Chien Wang[1*]    Euijoon Ahn[2]    Dagan Feng[1]    Jinman Kim[1]

[1] Biomedical and Multimedia Information Technology (BMIT) Research Group, School of Computer Science,

The University of Sydney, NSW 2006, Australia

[2] College of Science and Engineering, James Cook University, QLD 4811, Australia

**Abstract:** Over the last decade, supervised deep learning on manually annotated big data has been progressing significantly on computer vision tasks. But the application of deep learning in medical image analysis was limited by the scarcity of high-quality annotated medical imaging data. An emerging solution is self-supervised learning (SSL), among which contrastive SSL is the most successful approach to rivalling or outperforming supervised learning. This review investigates several state-of-the-art contrastive SSL algorithms originally on natural images as well as their adaptations for medical images, and concludes by discussing recent advances, current limitations, and future directions in applying contrastive SSL in the medical domain.

## 1 Introduction

In recent years, deep learning networks, such as convolutional neural networks (CNNs), have seen massive progress in image analysis techniques. LeCun et al.[1] showed that CNNs achieved superior performance on diverse computer vision tasks, including semantic segmentation, image classification, object detection, and activity recognition. When a large amount of data and manually annotated labels are available, CNNs can automatically learn to approximate the relationship between the data and its labels. This type of deep learning algorithm is called supervised learning[2]. However, supervised learning can also be limited by large-scale labelled image data availability, where manual annotation is costly, labour-intensive, time-consuming, and prone to human subjectivity and error[3, 4, 5]. CNNs have also been broadly applied with medical imaging modalities and are considered state-of-the-art in many medical image analysis applications[6], such as with breast cancer classification[7], COVID-19 detection[8] and skin lesion analysis[9].

A variety of methods have been proposed to deal with the problem of limited training images and labels. Transfer learning has become the established method for this problem. With transfer learning, the model is pre-trained on a larger image dataset, such as the ImageNet dataset of labelled natural images, and is then fine-tuned on a smaller dataset in the target domain that does not need to be from the same image domain, such as with a type of medical imaging modality[10]. Although transfer learning has demonstrated promising results in various medical imaging analysis applications[11, 12], there are known limitations[10, 13]. The primary limitation is that the image features extracted from the natural image dataset are not directly relevant to medical imaging datasets. Thus, supervised learning methods optimally designed using natural images do not necessarily translate well when applied to medical imaging analysis[10]. There are several key differences between medical images and natural images. As an example, medical images typically involve the identification of a small part of the images related to its pathologies or abnormalities, also known as regions of interest (ROIs), by utilizing variations in local textures from the whole image; examples of these are small red dots in retinal fundus images which are signs of microaneurysms and diabetic retinopathy[14], and white opaque local patches in chest X-ray images indicate consolidation and pneumonia. Natural image datasets, however, often have a large and salient object of interest in images. Another key difference is that, compared to natural images with diverse content and colours, a large variety of medical images, typically from X-ray, computer tomography (CT), and magnetic resonance imaging (MRI), are grayscale and have similar colours and content attributes across the image dataset, with fewer diversities and contrasts than natural images. Additionally, most medical image datasets have fewer image samples despite large variability in the image visual attributes between them, e.g., the number of images in the medical image datasets varying from one thousand[15] to one hundred thousand[16, 17]; in comparison, natural image datasets often have over 1 million images (e.g., ImageNet). Considering these differences between natural and medical images, transfer learning of natural image pre-trained model to medical image application is not always an effective solution. He et al.[18] demonstrated that pretraining on ImageNet merely accelerates the model

 

convergence early during the training process.

To address the scarcity of medical image labels, researchers have been using other deep learning methods that do not entirely rely on labelled image data, and instead utilize abundant unlabelled image data[19, 20]. To address these issues, Yann LeCun presented the first concept of self-supervised learning (SSL) in 2017. His talk at the AAAI 2020 conference[21] started to attract people's attention, and people gradually realized SSL had a potential future. He described, "In SSL, the system learns to predict part of its input from other parts of its input". SSL, as its name implies, creates supervisory information that is derived from the data itself. As represented in Fig. 1, there are some examples of SSL, such as predicting future data (yellow color) from past data (purple color) and predicting past data from present data (blue color). Take sequential datasets, for example, the target objects or images can be seen as anchors. The objects or images before these anchors can be seen as the past data, while the objects or images after these anchors can be seen as the future data. SSL has been widely employed in computer vision applications using natural images. For example, the Bootstrap Your Own Latent (BYOL)[22] method obtained better image classification results than some supervised learning approaches on the ImageNet dataset. Other experiments [23, 24] further demonstrated how SSL could efficiently learn generalizable visual representations from the images. For example, Tendle and Hasan[25] analyzed the SSL representations that were trained from the ImageNet source dataset and then fine-tuned on two different target datasets: one that were considerably different from the source dataset, and the other that was similar to the source dataset. By investigating the invariance property of learned representations, such as rotation, scale change, translation (vertical and horizontal) and background change, their experiments demonstrated that SSL representations produced better generalizability in contrast to supervised learning representations

Among SSL methods, contrastive self-supervised learning, or contrastive SSL, is the most successful approach that achieved outstanding performance close to, or even surpassing, the supervised learning counterparts [26]. Contrastive learning encourages learning feature representation with inter-class separability and intra-class compactness, which can assist in classifier learning [3, 27]. More specifically, intra-class compactness refers to how closely image representations from the same class are related to one another, and inter-class separability refers to how widely apart image representations are from different classes; this is due to SSL capability to learn without labels and therefore being able to leverage large datasets. Contrastive SSL has already been widely studied among both natural and medical image domains. There were several comprehensive reviews on natural images, such as contrastive learning of visual representations [28], generative learning and contrastive learning[3], pre-trained language models[29], and self-supervised contrastive learning[30]. However, these reviews did not focus on medical images that are different from natural ones with inherent medical image specific challenges and requirements. In addition, there were some SSL reviews on medical images [31,32]. Some of them discussed three categories, including predictive, generative, and contrastive learning, but in the contrastive learning category, the authors did not divide it into subsections and provide structured portioning of the work. However, our paper exclusively focused on predictive and contrastive learning and used subsections to describe more details of the related backgrounds. In this study, we provide a state-of-the-art review of SSL research, focusing on predictive learning and contrastive SSL learning, and their adaptation and optimization for the medical imaging domain. With the focus of our paper on medical images, where possible, we have used medical images in our example figures. Our contributions are as follows: Section 2 introduces a systematic categorization of the state-of-the-art predictive learning and contrastive SSL methods and discusses their methodology; these methods are based on natural images; Section 3 presents a review of predictive learning and contrastive SSL methods applied to medical images and their unique adaptations from the natural image method counterparts. Section 4 concludes the review and discusses the limitation of predictive learning and contrastive SSL on medical images and makes suggestions for future research and directions.

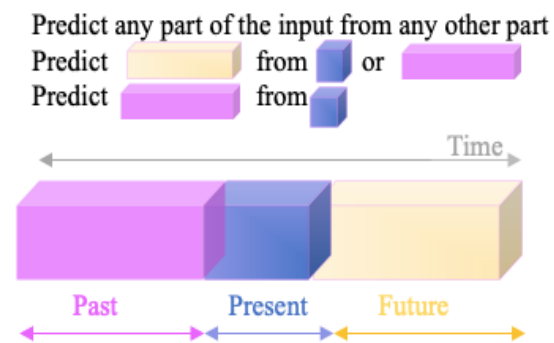

Fig. 1 The concept of self-supervised learning [1].

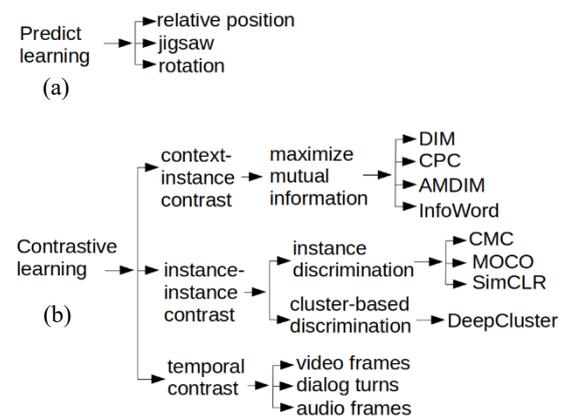

Fig. 2 Categorization of predictive learning(a) and categorization of contrastive SSL(b).

# 2 Predictive learning and Contrastive self-supervised learning (SSL)

## 2.1 Predictive learning

Through predicting geometric transformations of images, predictive learning tasks learn the structural and contextual semantics. Three types of spatially relevant position pretext tasks, as shown in Fig. 2(a), were described in this section: relative position, solving jigsaw puzzle, and rotation.

### 2.1.1 Relative position

The relative position model[33] was trained to learn the relationships between a selected patch and the patches around it. The relative position model selected a particular size of the area of an image sample and divides this area into certain number of disconnected patches. The number and the area in a patch, as shown in Fig. 3, were used for learning the relationship between the centre patch, called the anchor, and the neighbouring patches. As a result, the model learned the relationships between the patches. It was worth noting that the gaps between patches and the random displacement of patches prevent the model from learning the shortcut. Such a shortcut might be provided by low-level indications like boundary patterns or textures that continue between patches. There were three disadvantages with the relative position approach. First, multiple different objects could be included in two individual patches. For example, one patch contained the left atrium and another one consisted of the right atrium. There was no relevance between these two objects that are only located in the individual patches. As a result, no information could be learned about the relationship between those two objects. Second, in the relative position approach, CNNs could learn trivial features, such as the shared corners or edges of patches, instead of semantic feature representations that are beneficial to downstream discriminative tasks, including segmentation and classification tasks. Although some methods, such as the randomly jittering patches, were designed to prevent the model from learning trivial features, there are possibilities that patch positions would be learned from other places, such as background patterns. Third, since the relative position approach only involves the patches, it did not include the global information of images. This leaded to limited performance on downstream tasks that rely on global information of images, such as in image classification tasks. However, some of these tasks counted on ad hoc heuristics that might restrict the transferability and generalization of learned feature representations for the following downstream tasks.

### 2.1.2 Solving Jigsaw puzzle

One additional type of relative position was termed as "solving the jigsaw puzzle" [34]. The principal idea of this pretext task was to learn positional relations among divided patches of an input sample. In this approach, by solving the jigsaw puzzles, the algorithm learned to recognize the elemental structure of the objects, including objects and their relative parts. As shown in Fig. 4, within an image sample, the jigsaw puzzle solution first selected a particular size of the area that was relevant to the topic of interest. Then, this area was divided into nine puzzle patches shuffled based on the nine predefined permutation set as inputs. The model was trained to learn feature representation by correcting the order of those nine patches. The sequence of nine patches was used for the training model. The greatest challenge of the jigsaw puzzle was that the model required greater computational complexity and memory consumption. Noroozi et al.[34] also extended this to more complicated pretext tasks, such as the setting of 64 predefined permutations, demonstrating that more information on relative position can be learned.

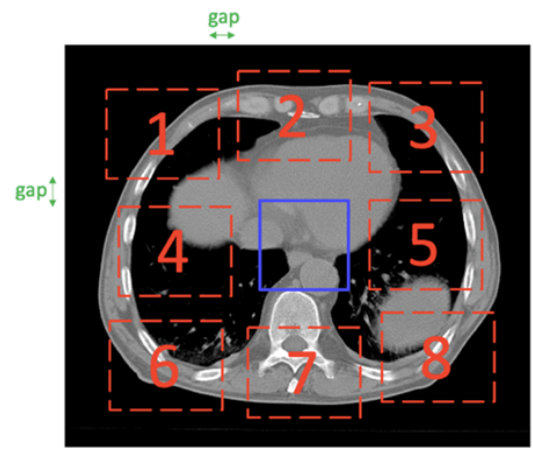

Fig. 3 An example of the predicting relative spatial position [33] pretext task on a CT lung image. The algorithm is trained to learn the relationships between a selected patch (blue centre) and the patches around it (red numbered patches).

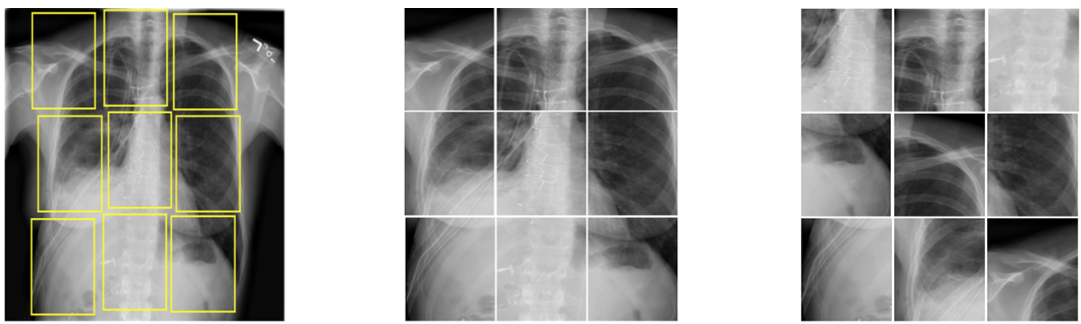

Fig. 4 An example of the "solving the jigsaw puzzle"[34] pretext task on an X-ray pneumothorax image.
The algorithm is trained to learn the positional relations among nine divided patches (yellow-framed patches).

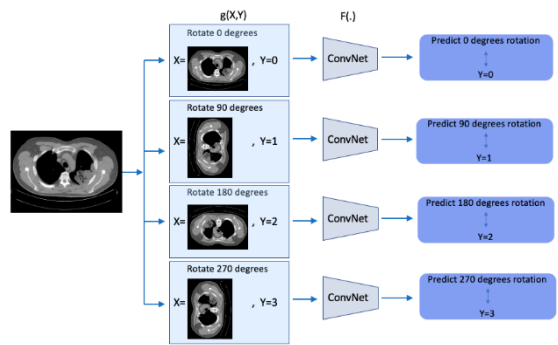

Fig. 5 An example of the predicting image rotations [35] pretext task on a CT lung image. The algorithm utilizes the rotation angle as a kind of supervision for training the model.

### 2.1.3 Rotation

Another context-based pretext task was designed for learning high-level semantic features by training the model to predict the degrees to which the input images were rotated. The rotation angle could be seen as a pseudo label for training the model. This was exemplified in Fig. 5. The result of [35] showed that the CT lung images rotated by angles of 0, 90, 180, or 270 degrees learn better feature representations than the other degrees rotations. Li et al.[36] also conducted research based on the rotation pretext task in which the angle was an expansion to 360 degrees. Lee et al.[37] trained the model with multiple pretext task learning strategies, including two types of transformations, rotations, and colour permutation, as those various self-supervised data augmentations enabled the reduction of the effects from the transformation invariant.

## 2.2 Contrastive self-supervised learning (SSL)

Contrastive learning is a method to learn feature representations via contrastive loss functions to distinguish between negative and positive image samples. Positive image samples are an augmentation of a target image (also called an anchor) while negative image samples are from other non-target samples within the training set. The contrastive learning approach encourages models to learn general-purpose feature representations that can be reused to enhance learning specifically in downstream tasks, e.g., segmentation and classification tasks, where the models are built using the learned features[38].

Contrastive learning methods typically vary in how they use unlabelled data to create or define negative and positive image pairs, and also in how they are sampled during training. Based on the idea of Liu et al.[3], contrastive learning categories are divided into two subcategories: context-instance contrast and instance-instance contrast. The context-instance contrast, also known as the global-local contrast, is concerned with modelling the relationship between a sample's local feature and its global context representation. Instance-instance contrast investigates the connections between the instance-level local representations of distinct samples. However, these two categories do not cater for the specific needs of sequential image or time series datasets. Any data that has elements that are arranged in sequences is referred to as sequential data[39]. Sequences of user actions, time series, and DNA sequences are a few examples. Yue et al.[40] mentioned that time-series medical images include rich spatial and temporal information. Therefore, we suggest a third category named temporal contrast, which is related to SSL designed for the sequential datasets. The three categorisations of contrastive SSL are shown in Fig. 2(b).

To train on unlabelled data, SSL uses "pretext" tasks as an alternative way to extract useful latent representations. Through solving the pretext tasks, pseudo labels, as supervisory signals, are generated automatically based on the dataset's properties. For example, with the rotation pretext task, the supervisory signals of "rotation angles", are derived from the unlabelled input samples. There are two different application paradigms for downstream tasks using the pretext task results. Fig. 6(a) shows that the first paradigm is learning transferable features. After solving the pretext tasks, the model will try to learn feature representation which can then be further trained for, e.g., fine-tuning for different tasks such as classification and detection. In contrast, Fig. 6(b) illustrates an example of learning "applicable embeddings" that refers to the pretext tasks used to directly learn generalizable features for downstream tasks.

Various pretext tasks are designed with those different augmentation transformations to capture the expected semantic or structural characteristics of images for downstream tasks. Before diving into subcategories, contrastive learning loss function is defined in Section 2.2.1 for a fundamental understanding of the SSL. Then, context-instance contrast learning and instance-instance contrast learning are described in Sections 2.2.2 and 2.2.3, respectively. Finally, temporal contrast is introduced in Section 2.4.

### 2.2.1 Contrastive learning

To learn meaningful features from the images, SSL use

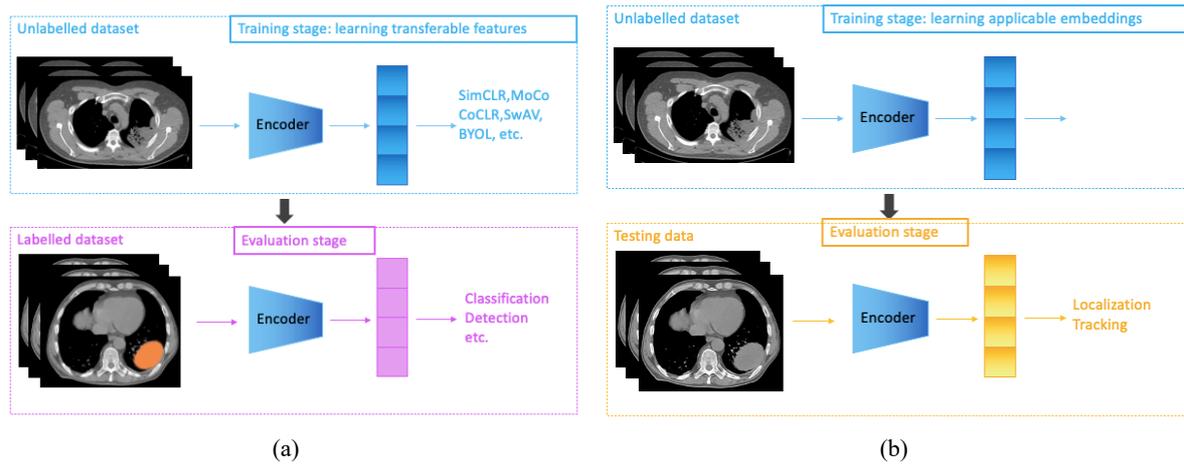

(a)                                                                                      (b)

**Fig. 6** Two different application paradigms for downstream tasks. In (a), further training such as fine-tuning is needed; while in (b), no annotation is needed for downstream tasks.

"data augmentation" techniques to generate additional data by increasing the diversity of the data transformation. Data augmentation involved image manipulation techniques, i.e., image scaling, cropping, flipping, padding, rotation, translation, and colour augmentation, such as brightness, contrast, saturation, and hue. The fundamental concept of contrastive learning was to group the images with its augmentations closer together and place the other images further away. This description can be expressed as:

$$score\big(f(x), f(x^+)\big) > score(f(x), f(x^-)) \quad (1)$$

where $f(x)$ is an encoder. The target image (also called an anchor), $x$, and the anchor's augmented sample, $x^+$, can be grouped as a positive pair. However, the anchor and other sample from the training dataset, $x^-$, are grouped as a negative pair. As a result, it can show that the score of the similar sample, $x$ and $x^+$, is higher than that of the dissimilar samples, $x$ and $x^-$. This score is a metric that compares the similarity between the two samples. Based on this concept, the following subsections discuss several common loss functions used in SSL.

### 2.2.1.1 Triplet loss

Triplet loss[41] was a type of metric learning with a similar concept to Equation 1, with changes in how it calculates the distance on the embedding space. In detail, minimizing the triplet loss, as in Equation 2, encourages the distance between the anchor and the positive sample to 0; and the distance between the anchor and the negative sample to be greater than the distance between the anchor and the positive sample plus with margin. When the representations created for a negative pair are distant enough, the purpose of the margin is to prevent effort wasted on enlarging this distance.

$$\mathcal{L} = \max\big(d(x, x^+) - d(x, x^-) + margin, 0\big) \quad (2)$$

Here, $d(x, x^+)$ denotes the distance between the anchor and the positive sample, $d(x, x^-)$ represents the distance between the anchor and the negative sample. The margin parameter is set to represent the minimum offset between the distances of the pairs.

### 2.2.1.2 Noise-contrastive estimation (NCE) loss[42] and InfoNCE loss[43]

To decrease the complexity of optimization, NCE was introduced to transform the calculation from multiclass classification problems to a binary logistic regression to classify data from noises. Inspired by NCE loss, InfoNCE loss used categorical cross-entropy loss to find positive samples from a collection of unrelated noisy samples. InfoNCE used similar data pattern for training, including one positive sample and many negative samples. However, InfoNCE loss often generated higher accuracies due to the selection of negative samples. This was explained by the grouping of the negative samples in the NCE algorithm as a unit for calculating an approximate value, while with InfoNCE, it calculated the negative samples as an individual sample and hence can keep more information about each of the data point. InfoNCE is formulated as:

$$L_N^{InfoNCE} = -\frac{E}{X}\left[\log\frac{f_k(x_{t+k}, c_t)}{\sum_{x_j \in X} f_k(x_j, c_t)}\right] \quad (3)$$

where $f_k()$ represents the density ratio, $t + k$ denotes the future time steps after $t$ from the dataset, $\{x_{t-k} ..., x_{t-1}, x_t ... x_{t+k}\} \in X$, where $f_k(x_{t+k}, c_t)$, and $f_k(x_j, c_t)$ can be seen as the positive sample pair and negative sample pair, respectively, in the collection of samples, $c_t$.

### 2.2.1.3 Mutual information (MI)

Mutual information[44] is a concept of reducing uncertainty about one random sample after observing another sample. Simply put, MI is a measure for assessing the relationship between arbitrary variables[45]. There were some MI applications, for example, Linsker et al.[46] which presented the InfoMax principle by using MI to calculate the relationship between the input and the output in the existence of processing noise. The relationship between InfoNCE and MI has been used in many state-of-art contrastive learning methods and after optimizing Equation 3, it can be expressed as:

$$I(x_{t+k}, C_t) \geq \log(N) - \mathcal{L}_N^{opt} \qquad (4)$$

where MI, $I(x_{t+k}, C_t)$, is equal to or larger than $\log(N)$, and $N$ is the number of samples, minus the optimized InfoNCE, $\mathcal{L}_N^{opt}$.

## 2.2.2    Context-instance contrast learning

Spatial context from images could be used to learn feature representations. It was originally from the concept of skip-gram Word2Vec[47] algorithm used in natural language processing (NLP), and later implemented for images by Doersch et al.[33] With spatial context, feature representations were learned by predicting the position of an image patch relative to other patches. The context-instance contrast learned the relationship between local and global image features. The idea of context-instance contrast was to capture the local features that can adequately represent the global features. In this category, the most popular algorithm is maximizing MI.

### 2.2.2.1 Maximizing MI

Unsupervised learning of feature representations could be achieved by maximizing MI between an input image and the output that was encoded by a deep neural network. The principle of high MI captures useful information rather than low-level noise. Tschannen et al.[44] conducted research on MI maximization for unsupervised or self-supervised representation learning, including Deep InfoMax (DIM)[48], Contrastive Multiview Coding (CMC)[49], and Contrastive Predictive Coding (CPC)[43].

### 2.2.2.1.1 Deep InfoMax (DIM)[48] and Augmented Multiscale DIM (AMDIM)[50]

Hjelm et al. [48] showed that, depending on the downstream task, it is often insufficient to learn effective representations by maximizing the MI between the encoder output (i.e., global MI) and the entire input. It is because global MI maximizes MI between global representation pairs, which included an entire image together with a single feature vector summarized from patches divided from the results of encoding input images. However, global Infomax has the problem that the model captured undesirable information such as trivial noise that was particular to local patches or pixels and that was useless for certain tasks such as image classification. This was because grabbing feature information particular to only belonging parts of the input through encoders did not enhance the MI with other patches that did not include those trivial noise. Hence, this issue arose the idea of local Infomax to encourage the encoders to learn feature representation that is shared across the patches of an input image. Hjelm et al.[48] showed that adding location information of the input into the object enables to considerably increase a representation's fitness for subsequent tasks. Hence, they proposed the ideas of global DIM and local DIM to train the encoders by maximizing MI between global and local patch features. Local Infomax maximizes MI between the summarized patch feature vector and each local patch feature, where both are extracted from different layers of the same structure of the convolutional network. Later, Bachman et al.[50] extended the idea of local DIM by maximizing MI between features generated through augmentation of each input image. The author improved the local DIM from the following three perspectives: data augmentation, multi-scale mutual information, and encoder. For data augmentation, they first performed a random horizontal flip and then some common data augmentations, including random in the crop, jitter in colour space, and grayscale transformation. This model learned features by maximizing MI between the global and augmented local features. To determine the similar part in augmented local features and global features. For multi-scale mutual information, the model learned features by maximizing MI within features from different layers with different scales. The MI between multi-scale features in the same images was higher than in different images. For the encoder, AMDIM improved the encoder based on the ResNet-base framework to control receptive files. The result is worse when there was too much overlap within the features of positive sample pairs.

### 2.2.2.1.2 Contrastive Predictive Coding (CPC)[43, 51]

Contrastive Predictive Coding[52, 53] focused on sequential data and utilizes useful information of previous sequential components of the data to predict the future sequential signal. During the predictive coding, the information of image content was embedded. Using autoregressive models, CPC encoded key shared information within different parts of the previous sequential signal to high-level latent space, and this was used to predict future that conditionally relies on the same shared information. This resulted in keeping a similar representation from the same images encoding more global and common features, and discarding low-level information and local variations, such as the noise. Additionally, the use of probabilistic contrastive loss for learning high-dimensional representations in latent embedding space maximized useful information for predicting future samples. Based on the ideas of NCE, CPC proposed InfoNCE and its relationship with MI. That is, minimizing the InfoNCE loss enabled maximizing a lower bound on MI between representations that were encoded.

## 2.2.3    Instance-instance contrast learning

Under instance contrast learning[54] category, instance comparisons were used from two points of view. The first was to design or modify contrastive loss functions and use specific structures for training SSL (see Section 2.3.1). The second was to directly compare instances to derive distinctive information within the instances (see Sections 2.3.2 and 2.3.3).

### 2.2.3.1 SSL design on contrastive loss function-based variation and specific structures

Within many strategies of designing SSL model, we discuss two ideas based on either the varied contrastive loss functions or specified structure in the subcategories.

#### 2.2.3.1.1 SSL design on contrastive loss function-based variation

When contrastive loss functions are designed or modified based on the principle of Equation 1, they had been applied to many different tasks for specified learning approaches. The five learning approaches introduced in this section are (1) multimodal learning, (2) local representation learning, (3) multi-scale learning, (4) texture representation learning, and (5) structural representation learning.

(1) For multimodal learning, most papers conducted SSL research on only one modality dataset. Hence, some studies have started working on multimodal SSL training to learn more meaningful semantic information that might compensate for each other. For computer vision, multimodality could group different types of resources, such as text and image, or different types of data formats, such as CT, X-ray, and MRI. (2) For local representation learning, most of the common instance-instance contrast learning methods concentrated only on extracting image-level global consistency between instances but neglect explicitly learning the distinctive local consistency within the instances. Distinctive local representations played a vital role in obtaining structural information for dense or per-pixel prediction tasks, including segmentation. (3) For multi-scale learning, some medical data were large, such as histology images. Such large images as input for training the network slowed down the calculation and increased the training time. Hence, for the domain of histopathology, some studies used relatively small areas or objects, such as nuclei, to predict whole histology images. However, some works utilized a variety of sizes of input for the training model and Yoo et al.[55] demonstrated how multi-scale local activations could enhance visual representation based on CNN activations. Finally, some SSL works designed the contrastive loss for learning (4) texture representation and (5) structural representation, respectively.

#### 2.2.3.1.2 SSL design on specific structures

Except for the design and modification of contrastive loss functions and the selected sample strategies, some works focused on the specific structures for training SSL, such as Siamese-based learning, and teacher-student-based learning. For the Siamese network learning, a Siamese neural network included two or more identical subnetworks which were used to estimate the similarity between two samples by two feature extractors with shared weights, and were utilized in many applications, such as the prediction of camera poses[56] and lip poses[57]. A large number of batch sizes or negative pairs applied in common SSL methods made them more difficultly be implemented on 3D medical datasets. Chen et al.[58] proved that the Siamese network could be used to avoid such problems on a 2D network. And, without relying on larger batch sizes or negative pairs, the Siamese network enabled to keep the spatial relationship in the embedding space through contrastive loss. For the Teacher-student-based learning, Teacher-student learning was a transfer learning approach in which the student network was taught by the teacher's network to predict the same result as the teacher's. A small network, the student network, could be learned by the labels produced by a complex model, the teacher network. Moreover, the Mean Teacher model, an extended model based on the teacher-student, was implemented for the medical image analysis tasks to average model weights to aggregate information after every step instead of every epoch. The Mean Teacher model also provided more robust intermediate representations since the weight averages captures all layer outputs, not just the top output.

### 2.2.3.2 Instance-based discrimination

There were a variety of techniques designed for collecting negative samples to compare with a positive sample in the training process, such as Memory Bank, Momentum Encoder PIRL[59], SimCLR[20], MoCo[19, 60, 61], and BYOL[22]. Though for different purposes, these methods could be considered to create dynamic dictionaries. In these dictionaries, the "queries" and "keys" were obtained from data, e.g., patches or images, which were embedding representations created through the query and key encoder networks, respectively. These encoders could be any CNNs[62]. SSL trained encoders to execute dictionary look-up: an encoded "query" should be comparable to its corresponding key while being distinct from others. The definition of query and key could be different. For example, Wu et al.[63] grouped a key and a query as a negative pair if they come from a different image and otherwise as a positive sample pair. However, Ye et al.[64] selected two random "views" of the same image using random data augmentation to create a positive pair. It is worth to notice that inconsistency was a big challenge in this method. Inconsistency existed between the query and key embedding representation. Specifically, inconsistency occurred when calculating the contrastive loss between the positive features from the query encoder that was updated each epoch and the negative features saved in the memory that was updated from several previous epochs. Hence, many approaches were proposed to solve this inconsistency. He et al.[19] hypothesized that it was possible to create consistent and large dictionaries during the training process and that in the dictionary, the keys should be represented through the similar or same encoder to provide consistency in comparisons to the query.

#### 2.2.3.2.1 Memory bank and Momentum encoder and Momentum Contrast (MoCo)[19]

Based on the principle of contrastive loss, the number of negative samples significantly affected the accuracy, which was proven by Nozawa et al.[65] In one batch, it included an original image, its augmented example, and many negative samples. The numbers of negatives sampled depended on the batch size and the large batch size means we could contain more negative samples. However, the batch size was limited by the GPU memory size. The memory bank was designed to address this problem by accumulating and regularly updating many embeddings of negative samples that resulted from the key encoder without increasing the batch size but with less gradient calculation from the encoded key query during training. Pretext-Invariant Representation Learning (PIRL) learned invariant representations by using a memory bank based on a pretext task related to solving the jigsaw puzzle.

Although memory banks could contain a larger number of negative samples, inconsistency existed between the query and key embedding representations that resulted from the query and key encoders, respectively. To address the inconsistency problem, MoCo decoupled the batch size from the negative samples by replacing the memory bank with a moving-averaged encoder called the momentum encoder. This momentum encoder was built as a dictionary-like queue that progressively replaced samples by enqueueing the current mini-batch and dequeuing the oldest mini-batch in this queue. The benefit of removing the oldest mini-batch that was outdated was to maintain consistency with the newest samples from the query encoder. By doing this, the number of negative samples could be increased without expanding the batch size. In brief, MoCo decreased the dependency on mini-batch size and utilized a momentum encoder to update the queue that involves previously processed samples to create contrastive pair encodings. This was defined as follows:

$$\theta_k \leftarrow m\theta_k + (1-m)\theta_q \qquad (5)$$

where the momentum coefficient, $m$, made the key encoder, $\theta_k$, slowly progress, driven by the query encoder, $\theta_q$, $(1-m)$. He et al.[19] proved that the performance was the best when m is 0.99 because this setting updated the key encoder slowly through a large part of the previous key encoders and a small part of the newest query encoder. This could keep a large and consistent dictionary that facilitates contrastive learning to train a visual representation encoder. Based on MoCo, the same team further designed MoCo v2[60] by adding an MLP projection head, data augmentation, and a cosine learning rate schedule.

#### 2.2.3.2.2 SimCLR[20]

SimCLR was an end-to-end learning architecture and learned feature representations by maximizing the agreement between dissimilar augmented views of the same input via a contrastive loss calculation[66]. Through experiments, the results of SimCLR showed four components that affect the quality of contrastive representation learning. The combination of data augmentation, random cropping, and colour distortion was shown to be better than other combinations or single transformations. Moreover, compared to supervised contrastive learning, unsupervised contrastive learning obtained greater advantages from longer training, larger batch sizes, and stronger data augmentation. However, similar to supervised learning, contrastive learning obtained an advantage from a deeper and wider framework. It is worth noticing that the introduction of the nonlinear projection head significantly improved the learning representations during training. Based on SimCLR, the same team further improved three steps for designing a semi-supervised learning framework called SimCLR v2[67].

#### 2.2.3.2.3 Contrastive Multiview Coding (CMC)[49]

Unlike DIM, CPC, and AMDIM using one view of the image, CMC worked on images that were acquired in more than one view. The goal of CMC was to learn feature representations with information shared between various sensory channels obtained from the same image. Specifically, CMC used NCE-based softmax cross-entropy loss to learn feature embeddings by maximizing MI between various views from the same scene. A 4-view dataset, NYU RGBD[68], from the same scene, was brought together in embedding space as positive samples, but the views from different scenes were pushed apart as the negative sample. CMC also proposed "core view" and "full graph" paradigms. The full graph outperforms not only because more cross-view learning can get better representation but also because full graph can deal with missing information of views.

#### 2.2.3.2.4 Bootstrap Your Own Latent (BYOL)[22]

Some contrastive learning methods in Section 2.3.2, such as SimCLR and MoCo, relied heavily on many negative samples for learning the discriminative features. Hence, those methods were sensitive to selecting data augmentation policies and require many trials to determine good data augmentation[69, 70]. Moreover, SimCLR required a long training time on large datasets, out of 3200 epochs on the 1.2 million ImageNet images[71], to obtain improved performance. Unlike SimCLR, BYOL used mean squared error (MSE) rather than a contrastive loss, so as to rely less on the availability of large-scale negative samples.

### 2.2.3.3 Cluster-based discrimination

In computer vision, the clustering algorithm was a class of unsupervised learning techniques that have been largely researched and applied. Although clustering techniques were the first stage of success in classifying images, relatively few papers introduced to apply them to CNNs end-to-end training on large scale datasets[72, 73]. A problem is that clustering techniques were primarily built on linear models for calculating the top of fixed features, and they seldom ever function when the features must be simultaneously learned.

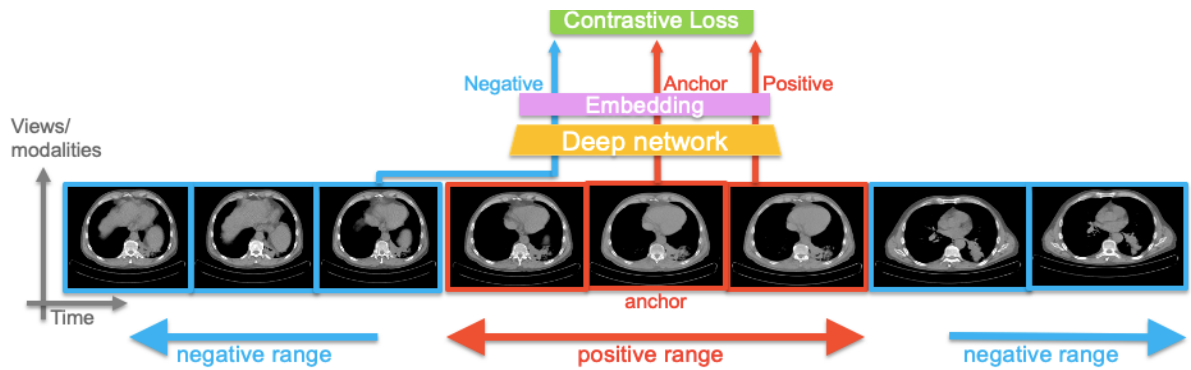

**Fig. 7** The selection of positive samples and negative samples from a set of adjacent frames.

Based on the clustering technique, DeepCluster was designed to simultaneously learn the features' cluster assignments and the neural network's parameters. More specifically, they iteratively clustered the features with a normal clustering algorithm, *k*-means, and utilized the cluster assignments as supervision signals to learn the parameters of the network. Unlike context instance contrast, clustering had the benefits of needing little domain knowledge and no particular signal from the inputs. In addition, some contrastive learning methods highly depended on the online calculation of many pairwise feature comparisons. Hence, the authors of SwAV[74] designed an online algorithm with a cluster-based idea to reduce the amount of computation. SwAV employed a "swapped" prediction technique in which the cluster assignments of one view were predicted based on the representation of another view. This method could work in large and small batch sizes without needing a momentum encoder or a large memory bank. A multi-crop technique was designed by making use of smaller-sized images to boost views without raising a training's memory or processing demands.

### 2.2.4 Temporal contrast

Medical imaging datasets, of CT or MR images, sometimes have follow-up scans with spatial or structural information. A sequence of CT or MR images, such as from left to right or from top to bottom of the patient's body, assists in learning more semantic representations. Compared to 2D data, videos or image sequences have richer information that allows to learn better feature representation through SSL. There are three common types of 3D SSL, including finding the similarity of adjacent frames, tracking the objects, and correcting the temporal order.

#### 2.2.4.1 Finding similarities of adjacent frames

First, adjacent frames should share similar features[75]. By training CNNs to learn the similarities within neighbourhood frames, contextual semantic representations could be learned. Moreover, temporal continuity[76] in sports activities, such as playing table tennis, and the characteristic of frames expressing a swing action should also be smooth. In this case, in the same sequence, the adjacent frames selected within a small design range were closer in embedding space than, frames selected from distant timesteps, as shown in **Fig. 7**. In addition to learning from the same video, Sermanet et al.[77] also learned from multi-view (multiple modalities) videos to obtain viewpoint and agent invariant feature representations. In this case, positive paired images obtained simultaneously with different viewpoints were closer in the embedding space than negative paired images obtained from a dissimilar time in the same sequence.

#### 2.2.4.2 Tracking the objects

Second, based on visual tracking-provided supervision for training models, Wang et al.[78] learned visual representations by unsupervised tracking within thousands of unlabelled moving videos. More specifically, two frames connected by a track should share a visual representation in feature space, such as cycling, because they probably corresponded to the same target of the moving object or were part of the object. Based on this idea, Walker et al.[79] utilized CNNs to learn similar objects that shared similar visual representations, and [80, 81] researched human poses. In this case[78], designed a ranking loss function to encourage, in feature space, the first two frames connected through a track to be much closer than the first frame and a random frame.

#### 2.2.4.3 Correcting for the temporal order

Third, it was a method to learn visual representation through an unsupervised sequential verification task, which corrected frame order from a sequence of video frames[82, 83, 84]. In this case, the correct order was a positive sample, and the wrong order was a negative sample, as shown in **Fig. 8**.

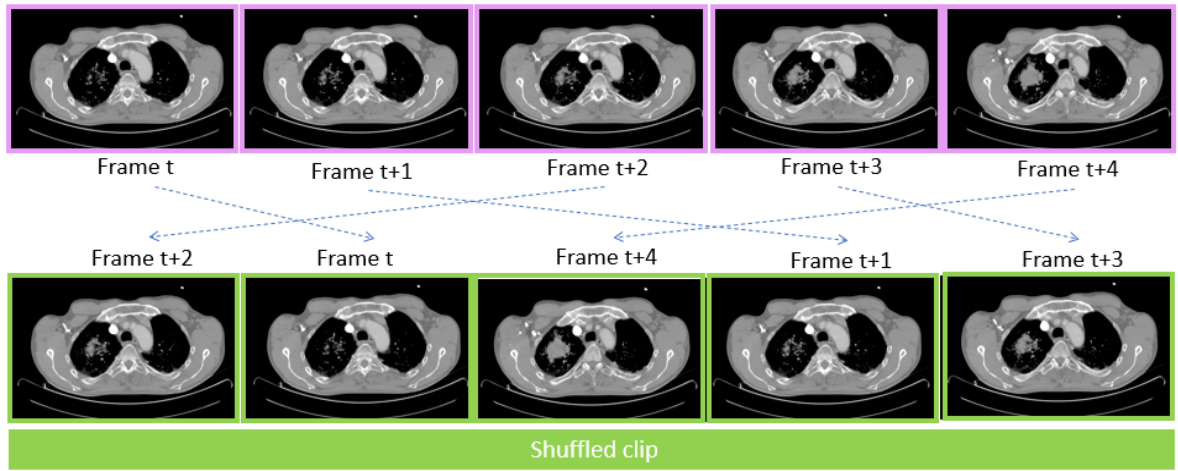

**Fig. 8** The positive example (correct order) and the negative example (incorrect order) from a sequence of video framesare trained to learn the semantic representations.

# 3 Predictive Contrastive SSL applied to medical images

Contrastive SSL has been broadly applied and optimized for medical images. There were four forms of contrastive SSL commonly applied to medical images: contrastive learning estimation, context-instance contrast learning, instance-instance contrast learning and temporal contrast SSL.

## 3.1 Predicting learning for medical image analysis

### 3.1.1 Relative position

SSL based on the relative position approach was also used in the medical area[85] for learning useful semantic features by utilizing image context restoration. Architecture with the combination of multiple SSL methods was used, including relative position prediction[33], colourization[86], exemplar CNNs[87], and inpainting[88]. In particular, the relative position was used to find the relationship between the central patch and eight nearby patches within a selected 3×3 selected patch grid. Inspired by the work of context prediction of adjacent patches[33], Blendowski et al.[89] proposed self-supervised 3D context feature learning, which included a new idea of image-intrinsic spatial offset relations with a heatmap regression loss. Jana et al.[90] used image context restoration[85] as the pretext task for checking non-alcoholic fatty liver disease that leaded to granular textural changes in the liver and could progress to liver cancer. Since one of the signs of non-alcoholic fatty liver disease was texture change in the liver, Chen et al.[85] encouraged the network to learn neighbouring pixel information for downstream tasks, including fibrosis and NAS score prediction. Based on [33], Li et al.[91] analysed the issue of COVID-19 severity assessment by training the SSL model to predict the relative location between two patches of the same CT slice. Fashi et al.[92] utilized the primary site information as pseudo-labels and modified the histopathology patch order for the training feature extractor. The added supervised contrastive learning loss boosted more robust feature representations for WSI classification.

### 3.1.2 Solving jigsaw puzzles

Based on solving jigsaw puzzles, SSL was applied to learn useful semantic features by blending patches from various medical imaging modalities[93]. This multimodal jigsaw task first drew random puzzle patches from dissimilar medical imaging modalities and combined them into the same puzzle. Combining these medical imaging modalities at the data level encouraged the model to derive modality-agnostic representations of the images and derive modality-invariant views of the objects, including tissues and organ structures. The learned feature representations from many medical imaging modalities could contain cross-modal information, which combined complementary information across the modalities. Taleb et al.[93] augmented multimodal data using cross-modal generation techniques to address modality imbalance problems in real-world clinical situations. In addition, their two modality experiments showed that the proposed multimodal puzzles learn powerful representations, even when the modalities were non-registered. One was on prostate segmentation of two MRI modalities, and another was liver segmentation of both CT and MRI modalities. By increasing performance on downstream tasks and data efficiency, it summed up that the multimodal jigsaw puzzle created better semantic representations when comparing the performance on each modality independently. Later, the same team proposed multimodal puzzle solving as a proxy task to assist feature representation learning from multiple image modalities[94]. Navarro et al.[95] compared and assessed the robustness and generalizability of both SSL and fully supervised learning networks on downstream tasks, including pneumonia detection in X-ray images and segmentation of various organs in CT images. By solving jigsaw puzzles on those medical datasets, they summarized that they efficiently learned feature mapping of object parts and their spatial arrangement through SSL. Based on the idea of a jigsaw puzzle-solving strategy, Manna et al.[96] learned spatial context-invariant features from magnetic resonance video clips to check knee medical conditions. They mentioned that the first work applied SSL to class imbalanced multilabel MR video datasets. Based on the jigsaw puzzles transformation[34], Li et al.[97] designed a self-supervised network by modifying two processes. The first was to increase the variety of permutations, and the second was to merge the jigsaw puzzles pretext task into the end-to-end semi-supervised framework. They applied the proposed semi supervised learning method to two medical image segmentation tasks, including nuclei[98, 99, 100] segmentation and skin lesion[101, 102, 103] segmentation. To classify cervix images as normal against cancerous, Chae et al.[104] presented a new patch of SSL based on puzzle pretext tasks to predict the relative position. Because they found that the pivotal area of the image to search for cervix cancer was highly potential around the centre and the irrelevant parts were near the periphery. In the domain of histopathology, based on the relative patch algorithm, Santilli et al.[105] implemented domain adaptation from the skin to breast spectra because of the low-level resemblance in the outline between skin tissue and breast cancer. They applied a relative patch pretext task for training on skin data to learn positional relations among divided patches of an input sample and then transferred the learned weights to the following downstream task, breast cancer classification. Zhuang et al.[106, 107] and Tao et al.[108], inspired by the jigsaw puzzle, proposed a novel 3D proxy task by playing a Rubik's cube, called the Rubik's cube recovery. Since the jigsaw puzzle was designed for 2D data, the Rubik's cube recovery was introduced for 3D volumetric data. During the Rubik's cube recovery process, rich feature information from 3D medical images was obtained, including cube rearrangement and cube rotation. This enforced the model to learn the features invariant from both translational and rotational perspectives. It is worth noting that the difficulty increased when the cube rotation operation was added to the Rubik's cube recovery, as it encouraged networks to exploit more spatial information. Li et al.[109] extended the Rubik's cube by adding a random masking operation for obtaining feature representations from the COVID-19 and negative CT volumes.

### 3.1.3 Rotation

Li et al.[110] observed that each fundus image included obvious structures, such as the optic disc and blood vessels, that were sensitive to orientations. Hence, they proposed a rotation-oriented collaborative approach to learning complementary information, including rotation-related and rotation-invariance features. With these two pretext tasks, vessel structures in fundus images and the discriminative features for retinal disease diagnosis were learnt. In addition to the rotation pretext task, Yang et al.[111] applied elastic transformation prediction[112], to cross-modality liver segmentation from CT to MR. Inspired by [35, 113, 114], Liu et al.[115] presented SSL based on a 3D feature pyramid network for assisting multi-scale pulmonary nodule detection. Dong et al.[116] classified focal liver lesions by utilizing several relative position pretext tasks, such as predicting the relative position between patches of an input, predicting the rotation, or solving a jigsaw puzzle. Imran et al.[117] presented a new semi-supervised multiple-task model utilizing self-supervision and adversarial training to classify and segment anatomical structures on spine X-ray images. Several pretext tasks were used several SSL simultaneously for medical imaging analysis, such as the studies that worked on the combination of rotation prediction[35] and jigsaw puzzle assembly[34]. However, Tajbakhsh et al.[118] combined two different types of SSL, such as a rotation (contrastive SSL) and reconstruction[119] and colorization[120] (generative SSL), on retinal images for diabetic retinopathy classification. In histopathology, Koohbanani et al.[121] utilized and combined various self-supervised tasks for domain-specific and domain agnostic purposes to obtain contextual, multiresolution, and semantic features in pathology images. Vats et al.[122] adopted those two pretext tasks for wireless capsule endoscopy diagnosis.

## 3.2 Contrastive learning estimation for medical image analysis

To focus on abnormalities, Liu et al.[123] introduced a learnable alignment module into contrastive learning to alter all input samples to be geometrically canonical. More specifically, after extracting high-level feature representations of the image pair, the highly structured character of inputs was used to calculate the L1 distance between corresponding pixels on the positive and negative images. The result could be seen as an indication of possible lesion location on the latter. Their model could alleviate the difference in scales, angles, and displacements of X-ray samples created under bad scan conditions. They demonstrated that the learned features represent localization information that enabled better identification and localization of downstream tasks, including infiltration, mass and pneumothorax diagnosis.

### 3.2.1 Contrastive learning

#### 3.2.1.1 Triplet loss for medical application

Xie et al.[124] proposed a novel SSL framework with scale-wise triplet loss and count ranking loss, to encourage neural network to automatically learn the information of nuclei quantity and size from the raw data for nuclei segmentation.

### 3.2.1.2 Noise-contrastive estimation (NCE) loss[42] and InfoNCE[43] for medical image analysis

Sun et al.[125] presented a context-aware self-supervised representation learning approach for learning anatomy-specific and subject-specific representations at the patch and graph levels, respectively. Interestingly, they utilized InfoNCE loss to learn patch-level textural features and contrastive learning objectives for learning graph-level representation. They also took advantage of MoCo, including a queue of data samples and a momentum update scheme to enhance the number of negative samples during training. The features learned through the proposed method demonstrated better performance in staging lung tissue abnormalities associated with COVID-19 than those learned by other unsupervised baselines, such as MedicalNet, Models Genesis, and MoCo. Most existing methods that used the maximization of MI as contrastive loss utilized image pairs for training; however, Zhang et al.[126] made use of image-text pairs. Their work enhanced visual representation learning of medical images by taking advantage of the combined information from textual data and image pairs. Through a bidirectional contrastive objective loss between those two different modalities, this approach depended on maximizing the agreement between real medical representation image-text pairs and randomly chosen pairs. More specifically, bidirectional contrastive objective losses were utilized similarly to the InfoNCE loss. Minimizing this loss encourages encoders to reserve the MI between real representation image-text pairs. Punn et al.[127] utilized the Barlow Twins framework to pre-train an encoder through redundancy reduction, similar to the InfoNCE objective, to learn feature representation over four biomedical imaging segmentation tasks, including cell nuclei, breast tumour, skin lesion, and brain tumour. Except for InfoNCE-based contrastive loss based on the MoCo framework, Kaku et al.[128] added additional two losses, mean squared error (MSE) and Barlow Twins (BT). By minimizing the MSE of feature representations between the intermediate layer or using BT to make their cross-correlation matrix closer to an identity matrix, the model was encouraged to learn augmentation-invariant feature representations that were not only focused on the final layer of the encoder but also extracting the intermediate layers. Their results showed performance was better than MoCo on three medical datasets, including breast cancer histopathology, NIH chest X-ray and diabetic retinopathy. Taher et al.[129] found instance-based objectives learned the most discriminative global feature representations, which might not be sufficient to discriminate medical images. Hence, inspired by the integration of generative and discriminative approaches, Preservational Contrastive Representation Learning (PCRL)[130], Taher et al.[129] developed an SSL framework, context-aware instance discrimination, to encourage instance discrimination learning with context-aware feature representations.

### 3.2.2    Context-instance contrast learning for medical image analysis

#### 3.2.2.1 Maximizing MI for medical image analysis

##### 3.2.2.1.1 Deep Infomax (DIM)[48] and Augmented Multiscale DIM (AMDIM)[50]

Chen et al.[131] combined two different types of self-supervised methods, one from the context-instance category, DIM, and another from the instance-instance category, SimCLR[20], for learning disease concept embedding. They utilized the proposed model to extract medical information from electronic health records and disease retrieval.

##### 3.2.2.1.2 Contrastive Predictive Coding (CPC)[43]

Stacke et al.[132] implemented and evaluated CPC on histopathology. After experimenting with some model and data-specific parameters on CPC models on histopathology images, those models were estimated for linear tumour classification on three tissue types. This work summarized the restriction of the learned representation for linear tumour classification on histopathology images because only low-level features in the first CPC layers were used. The diversity of distribution of the histology dataset makes little difference for linear tumour classification on histopathology images. Taleb et al.[133] extended this idea to a 3D CPC version. Instead of the time sequence dataset used in CPC, 3D CPC utilized a feature representation set obtained from patches cropped from the upper or left part of the 2D image sample to predict the encoded feature representations of the remaining part, lower or right part. In addition, they also developed a 3D version for rotation prediction, relative patch location, jigsaw puzzles, and exemplar networks. They demonstrated that the feature representations learned from 3D models were more accurate and efficient for solving downstream tasks than training the models from scratch and pretraining them on 2D slices. Zhu et al.[134] investigated the feature complementarity within multiple SSL approaches and presented a greedy algorithm to add multiple proxy tasks. More specifically, based on the assumption that a weaker correlation indicated a higher complementarity between two features, they calculated the correlation measure between the features created by different proxy tasks and then utilized the greedy algorithm to iteratively include a proxy task in the current task pool to form a multitask SSL framework. They applied it to the 3D medical volume brain haemorrhage dataset by adding multiple proxy tasks, including 3D rotation, Models Genesis[135], 3D CPC, and the Rubik's cube. After locating the potential lesions through super voxel estimation utilizing simple linear iterative clustering, Zhu et al.[136] calibrated CPC to learn 3D visual representation. More specifically, calibrating the CPC scheme on the sub volumes cropped from super voxels embedded the rich contextual lesion information into 3D neural networks. Cerebral haemorrhage classification and benign and malignant nodule classification were implemented using the proposed method on the brain haemorrhage and lung cancer datasets, respectively.

### 3.2.3 Instance-instance contrastive learning for medical image analysis

### 3.2.3.1 SSL design on contrastive loss function-based variation and specific structures for medical image analysis

3.2.3.1.1 SSL design on contrastive loss function-based variation

Based on the principle of the contrastive learning loss function, some papers worked on selecting positive and negative samples. For example, Jian et al.[137] combined a multi-layer network and VGG-16 to discriminate images with helicobacter pylori infection from images without helicobacter pylori infection well. However, some papers modified the principle of the contrastive learning loss function for particular applications, such as the following five applications. (1) Learning multimodality for medical applications—Holmberg et al.[138] proposed a new large-scale and cross-modality SSL in the field of ophthalmology. This SSL pretext task encoded shared information between two high-dimensional modalities, including infrared fundus photography and optical coherence tomography. The fundus representation learned from the SSL pretext task contains disease-relevant features that were efficient for downstream diabetic retinopathy classification and retinal thickness measurement. However, the audio and video data used for training SSL could be seen, e.g., in [139]. In detail, by assuming that there was dense correspondence between the ultrasound video and the relevant narrative diagnosis/interpretation speech audio of the sonographer, Jiao et al.[139] proposed SSL with multimodal input, including ultrasound video-speech raw data. Interestingly, to learn domain-agnostic feature representation, Tamkin et al.[140] designed the model architecture and objective to pretrain on six unlabelled datasets. Those datasets from various domains include text, natural images, medical imaging, multichannel sensor data, speech recordings and paired text and images. (2) Learning local representation for medical applications—Xie et al.[141] also focused on local regions by utilizing spatial transformation to create dissimilar augmented views of the same input. This encouraged consistent latent feature representations of the same region from different views of the same input image and assured such consistency by minimizing a local consistency loss. The proposed algorithm was for pretraining to initialize a downstream network and improve four publicly available CT datasets, including two tumours and 11 different types of primary human organs. Chaitanya et al.[142, 143] not only used global contrastive learning but also proposed a local version of contrastive learning. In particular, the local version of contrastive learning loss encouraged feature representations of local areas in an image to be similar with different transformations but dissimilar to different local areas in the same image. The combination of global and local contrastive learning benefited the downstream MRI segmentation task. One similar work proposed by Ouyang et al.[144, 145] employed super pixels pseudo labels and was devised for the tuning-free few-shot segmentation task, including cardiac segmentation of MRI dataset, and organ segmentation of abdominal MRI and CT dataset. Furthermore, the same team[146] designed a local pixel-wise contrastive loss to learn discriminative pixel-level feature representations. This enabled the model to learn better inter-class separability and intra-class compactness for the segmented classes on three public medical datasets with two anatomies, including cardiac and prostate. Yan et al.[147] proposed a pixel-level contrastive learning framework with a coarse-to-fine architecture to learn both local and global information and designed customized negative sampling strategies. More specifically, the global embedding was trained to discriminate various body parts on a coarse scale, assisting the local embedding to concentrate on a smaller region to distinguish finer features. The learned embeddings were applied in different downstream areas, such as landmark detection and lesion matching, on various radiological image modalities, including 3D CT and 2D X-ray of varying body parts, such as the chest, hand, and pelvis. (3) Learning multi-scale information for medical applications—in histopathology, Sahasrabudhe et al.[148] proposed a self-supervised method for nuclei segmentation on whole-slide histopathology images. They utilized scale classification as a self-supervision signal under the hypothesis that the texture and size of nuclei could be seen as the level of magnification at which a patch was obtained. Sun et al.[149] introduced a multi-scale SSL framework to precisely segment tissues for a multi-site paediatric brain MR dataset with motion/Gibbs artifacts. (4) Learning texture representation for medical applications—Chen et al.[150] proposed a new computer-aided diagnosis approach with contrastive texture learning loss to learn cervical optical coherence tomography images' texture features. (5) Learning structural representation for medical applications—Tang et al.[151] estimated the similarity between original and augmented images through the designed structural similarity loss for enhancing medical image classification.

3.2.3.1.2 SSL design on specific structures

Recently, Siamese network and Teacher-student were the popular structures applied in medical area. Siamese network learning for medical applications—Spitze et al.[152] utilized a Siamese network to calculate spatial distances between image patches sampled randomly from the cortex in random sections of the same brain. Learning to discriminate several cortical brain areas through their model implicitly indicated that the designed pretext task was suitable for high-resolution cytoarchitectonic mapping. Due to the benefits of decreasing the calculational expense of 3D medical imaging, Li et al.[153] extended a 2D Siamese network to a 3D Siamese network to avoid using negative pairs or large batch sizes. Their proposed SSL coped with an imbalance problem that assisted the learned radiomics features for two downstream classification tasks, including discrimination of the level of brain tumours on the MRI dataset and the stage of lung cancer on the CT dataset. Ye et al.[154] applied a Siamese network on stereo images for accessing depth in robotic surgery. For kidney segmentation from abdominal CT volumes, Dhere et al.[155] used a Siamese CNN to classify whether a given pair of kidneys belonged to the same side. They designed a proxy task by utilizing the anatomical asymmetry of kidneys, and the slight variation in shape, size, and spatial location between the left and right kidneys varied slightly. Moreover, some patients were scanned many times in a so-called longitudinal manner to track therapy or to estimate changes in the disease state. Hence, some studies on longitudinal information of the scans were used for training a Siamese network to compare the embeddings of scans from the same person or different persons. To pre-train on the example of T2-weighted sagittal lumbar MRIs, Jamaludin et al.[156] utilized SSL with a Siamese CNN trained through the two losses described as follows: (1) a contrastive loss on the pairs of images scanned from the same patient (i.e., longitudinal information) at different points in time and on the pairs of images of different patients, and (2) a classification loss was used to predict vertebral bodies' level and disc degeneration radiological grading. Rivail et al.[157] presented a self-supervised method based on a Siamese network for modelling disease progression from longitudinal data, such as longitudinal retinal optical coherence tomography. Taking advantage of a generic time-specific task, this self-supervised model learned to evaluate the time interval between pairs of scans obtained from the same patient. Teacher-student learning for medical applications—Li et al.[158] designed a new SSL approach based on the teacher-student architecture to learn distinguishing representations from gastric X-ray images for a downstream task, gastritis detection. One of the student-teacher frameworks, Mean Teacher[159], was integrated by Liu et al.[160] in the pretraining process for semi-supervised fine-tuning for thorax disease multilabel classification. Park et al.[161] used information distillation between teacher and student framework and the vision transformer model for chest X-ray diagnosis, including tuberculosis, pneumothorax, and COVID-19. You et al.[162,163] also demonstrated the distillation framework improved on medical image synthesis, registration and enhancement on the Left Atrial Segmentation Challenge (LA) and the NIH pancreas CT dataset. Later, they also proposed another semi-supervised approach that used stronger data augmentation and understood the nearest neighbours whose anatomical characteristics were homogeneous from the same class but distinct for other classes in unlabelled and clinically unbalanced circumstances [164].

### 3.2.3.2 Instance-based discrimination for medical image analysis

#### 3.2.3.2.1 Memory bank Momentum encoder and Momentum Contrast (MoCo)[19]

The model[165] that incorporated PIRL and transfer learning could learn the invariance property for skin lesion analysis and the results outperformed those obtained only using transfer learning or only using SSL. Taking advantage of MoCo while reducing dependency on batch size, Sowrirajan et al.[166] utilized it as a fundamental framework for reducing two constraints caused during training on the X-ray image. These two constraints were large X-ray image sizes and high computational requirements. The proposed MoCo-CXR model that adjusted the data augmentation strategy used in MoCo obtained high-quality feature representations and transferable initializations for the following detection of pathologies on chest X-ray images and across different chest X-ray datasets.

Several works used MoCo for COVID-19 diagnosis. Sriram et al.[167] applied MoCo to the COVID-19 adverse event prediction task from both single and multiple images and oxygen requirements prediction. To learn meaningful and unbiased visual representations for decreeing the risk of overfitting, He et al.[168] integrated contrastive SSL training on a similar dataset into transfer learning. Zhu et al.[169] utilized the combination of rotation and division as the supervisory signal on the SSL framework for COVID-19 classification on the small shot scenario. Based on the MoCo v2 algorithm, hierarchical pretraining, applied by Reed et al.[170], consistently converged to learn representations for experimenting on 15 of the 16 diverse datasets, spanning visual domains, including medical, driving, aerial, and simulated images. For medical datasets, they checked whether any of the five conditions were in each image of the CheXpert dataset[171] and classified 4-way pneumonia on the Chest-X-ray-kids dataset[172]. Hierarchical retraining was a way to train models on datasets that were gradually more similar to the target dataset. Liang et al.[173] also employed MoCo v2 as the base for conducting a neural architecture search to search for an optimal local architecture from its data. They applied it to CheXpert-14[171] and ModelNet40[174] for five classification tasks, including pleural effusion, atelectasis, consolidation, edema, and cardiomegaly. Interestingly, to train the encoder that could extract feature representation from the panoramic radiograph of the jaw, Hu et al.[175] utilized MoCo v2 to train the feature extractor on massive healthy samples. The Joint with localization consistency loss and patch-covering data augmentation strategy could improve the model's reliability. Wu et al.[176,177] integrated contrastive learning with federated learning [178, 179, 180] to collaboratively learn a shared image-level representation. Federated learning trained an algorithm within different decentralized edge devices to learn a shared model and each device kept local data samples without exchanging them. They experimented on 3D cardiac MRI images using MoCo architecture for local contrastive learning. Dong et al.[181] also federated SSL based on MoCO for COVID-19 detection. He et al.[182] combined a new surrogate loss proposed by Yuan et al.[183] with MoCo-based SSL for computer-aided screening of COVID-19 infected patients utilizing radiography images. This novel surrogate loss maximised the area under the receiver operating characteristic curve (AUC), and this combination facilitated vital metrics while also keeping model trust. Saillard et al.[184] implemented MoCo v2 on histology images from The Cancer Genome Atlas dataset for microsatellite instability prediction in gastric and colorectal cancers. Tomar et al.[185] applied a Style Encoder to the SSL framework utilizing volumetric contrastive loss through Momentum Contrast[19]. Style Encoder was designed to encourage content-invariant image-level feature representation that gathered similar styled images and dispersed dissimilar styled images.

### 3.2.3.2.2 SimCLR[20]

Azizi et al.[186] proposed a new method, Multi-Instance Contrastive Learning (MICLe), to classify two kinds of medical images, dermatology on camera images and multilabel on chest X-ray images. Unlike the traditional pretrained model, this work pretrained to the model on unlabelled ImageNet using SimCLR. Then, this work used MICLe to perform self-supervised pretraining on unlabelled medical images to create moderate positive pairs. Finally, supervised fine-tuning was performed on labelled medical images. Gazda et al.[187] proposed a self-supervised deep neural network that combined SimCLR and MoCo to first pretrain on an unlabelled CheXpert dataset of chest X-ray images and then transferred the pretrained representations to downstream tasks, including COVID-19 and pneumonia detection tasks, that is, the classification of respiratory diseases. In the histopathology domain, based on SimCLR, Ciga et al.[188] discovered that the combination of multiple multiorgan datasets with several types of staining and resolution properties enhanced the quality of the learned features. Li et al.[189] addressed whole-slide image classification by training the feature extractor SimCLR. Interestingly, for SimCLR training, they used patches as inputs extracted from the whole slide image and were densely cropped without overlap, which could be seen as an individual input. Ciga et al.[190] also implemented SimCLR for breast cancer detection in histopathology. Mojab et al.[191] verified the proposed model, a SimCLR-based framework with transfer learning, on real-world ophthalmic imaging datasets for glaucoma detection. Schirris et al.[192] utilized a SimCLR-based feature extractor pre-trained on histopathology tiles and extended DeepMIL[193] classification framework for Homologous Recombination Deficiency (HRD) and Microsatellite Instability (MSI) classification on colorectal and cancer dataset. Zhao et al.[194] added the Fast Mixed Hard Negative Sample Strategy to rapidly synthesise more hard negative samples[195] through a convex combination for training. The proposed model was pre-trained in a self-supervised way on the Chest X-ray of Pneumonia dataset and fine-tuned in a supervised way on the COVID-CT dataset. Wicaksono et al.[196] combined two types of contrasting learning, rotation, and jigsaw puzzle from context contrastive instance category and SimCLR v1 from instance contrastive learning, for the human embryo image classification task. Based on SimCLR, Manna et al.[197] also proposed the asymptotic study of the lower bound of the designed novel loss function to test the MRNet dataset, which composed magnetic resonance videos of the human knee. You et al.[198] presented two learning strategies for the volumetric medical image segmentation task. One used a voxel-to-volume contrastive algorithm to obtain global information from 3D images, and the other used local voxel-to-voxel distillation to better utilize local signals in the embedding space. Yao et al.[199] were motivated by contrastive learning[20, 200], which localized the object landmark with only one labelled image available in a coarse-to-fine fashion to create pseudo-annotation for training a terminal landmark detector. The proposed model demonstrated the high-performance cephalometric landmark detection, comparable to the popular fully supervised approaches utilizing more than one training image. Ali et al.[201] used 3D SimCLR during pretraining and the Monte Carlo dropout during prediction on two tasks, including 3D CT pancreas tumour and 3D MRI brain tumour segmentation. Inglese et al.[202] followed a similar optimization method of SimCLR to train an SSL network for distinguishing between two diagnostically different systemic lupus erythematosus patient groups. To learn task-agnostic properties, such as texture and intensity distribution, from heterogeneous data, Zheng et al.[203] first aggregated a dataset from various medical challenges. Then, they presented hierarchical SSL based on SimCLR with contrasting and classification strategies to provide supervision signals for image-, task-, and group-level pretext tasks. On the downstream tasks, they segmented the heart, prostate, and knee on the MRI dataset and the liver, pancreas, and spleen on the CT dataset.

### 3.2.3.3 Cluster-based discrimination for medical application

Abbas et al.[204] proposed a new SSL mechanism, 4S-DT, assisted coarse-to-fine transfer learning according to a self-supervised sample decomposition of unannotated chest X-ray input. Super sample decomposition[205] was a pretext task that trained networks using cluster assignments as pseudo labels. The coarse transfer learning utilized an ImageNet pre-trained CNN model for classifying pseudo-labelled chest X-ray images, creating chest X-ray related convolutional features. Fine transfer learning was used in downstream training tasks from the chest X-ray recognition tasks to COVID-19 detection tasks. In histopathology, Abbet et al.[206] conducted research on learning cancerous tissue areas that could be utilized to enhance prognostic stratification for colorectal cancer. They presented an SSL method that combined the learning of tissue region representations and a clustering metric to extract their underlying patterns. Mahapatra et al.[207] utilized one of the deep clustering methods[208], named SwAV, without using class attribute vectors commonly used for natural images. They proved the effectiveness of the proposed model across different datasets with at least three disease classes. Chaves et al.[209] evaluated five SSL methods, including InfoMin, MoCo, SimCLR, BYOL, and SwAV, for diagnosing skin lesions; they compared those SSL methods and three self-supervised pipelines on five test datasets with in-distribution and out-distribution scenarios. They summarized that self-supervision is competitive both in increasing accuracies and decreasing outcomes' variability. Chen et al.[210] developed an SSL strategy to perform joint deep embedding and cluster assignment for dMRI tractography white matter fiber clustering. Ciga et al.[211] utilized a two-step pretraining on three popular contrastive techniques, SimCLR, BYOL and SWaV, to validate better performance on two natural and three medical images, including ChestX-ray8, breast ultrasound, and brain tumour MRI. Islam et al.[212] pretrained and compared models within fourteen different SSL approaches for pulmonary embolism classification on CT pulmonary angiography scans.

### 3.2.4 Temporal contrastive SSL for medical image analysis

Temporal contrastive SSL learned feature representation by grabbing the spatial or structural information between adjacent frames. Sequential images utilized two kinds of a way as self-supervision for the training model, such as the objects shown in the adjacent frames or the process of correcting frame order.

### 3.2.4.1 Finding similarities of adjacent frames for medical image analysis

One of the most common applications of temporal contrastive SSL was with finding the similarity in adjacent frames. This enabled the mode to learn contextual semantic representations. In histopathology, Gildenblat et al.[213] utilized the image characteristic that spatially adjacent histopathological tissue image slices were more similar to one another than distance slices, which was used to train on a Siamese network for learning image similarity. In another application, due to the cardiac MR scans composed of different angulated planes relative to the heart, Bai et al.[214] learned feature representation, through the proposed model, from information automatically defined by the heart chamber view planes. That information included anatomical positions and the relative orientation of long-axis and short-axis views could be used to create a pretext task for SSL training. Kragh et al.[215] implemented a self-supervised video alignment method, temporal cycle consistency[216], to obtain temporal similarities between embryo stores, and this information to predict pregnancy possibility. By utilizing the position information in volumetric medical slices, Zeng et al.[217] provides a new position contrastive learning framework to produce contrastive data pairs. The framework can successfully get rid of false negative pairings in the currently common contrastive learning techniques for medical segmentation.

### 3.2.4.2 Tracking objects for medical image analysis

Lu et al. [218, 219] designed a pretext task to predict the density map of fibre streamlines that were the representations of generic white matter pathways for white matter tracts. They took advantage of two characteristics of the fibre streamlines. These fibre streamlines could be calculated with fibre tracking obtained automatically with tractography, and the density map of fibre streamlines was acquired as the number of streamlines cross each voxel. In short, fibre streamlines were jointed line segments with directions and could be seen as white matter pathways that provide supervision. To segment white matter tracts on diffusion magnetic resonance imaging scans, learned features of white matter tracts through the designed pretext task could predict the density map of fibre streamlines from the training data obtained through tractography.

### 3.2.4.3 Correcting frame orders from 3D medical images

The process of correcting frame orders from shuffled frames assisted the model in learning feature representation. Zhang et al.[220] utilized spatial context information in 3D CT and MR volumes as a source of supervision by solving the tasks of transversal 2D slice ordering for fine-grained body part recognition. Nguyen et al.[221] also demonstrated that predicting the 2D slice order in a sequence could obtain both spatial and semantic features for downstream tasks, the detection of organ segmentation, and intracranial hemorrhage. Jiao et al.[222] corrected the order of a reshuffled fetal ultrasound video. By utilizing the characteristics of the tube-like structure of axons, Klinghoffer et al.[223] learned feature representation by training the model to predict the permutation that was utilized to reformulate the slices of each input 3D microscopy subvolume for axon segmentation. The design of the pretext task, resolution sequence prediction[224], was inspired by the approach in which a pathologist looked for cancerous regions in whole-slide images. More specifically, a pathologist zoomed in and out several times to inspect the tissue at high to low resolution to acquire the details of individual cells and the surrounding area. Srinidhi et al.[224] utilized multiresolution contextual information as a supervisory signal to train a designed SSL network. This network learned visual representations by predicting the order of sequences of resolution that could be generated from the multiresolution histology whole slide image patches.

# 4 Conclusions and future directions

This study reviews the state-of-the-art contrastive SSL algorithms on natural images, along with their novel adaptations for medical imaging data. We cover fundamental problems when implementing SSL in medical areas and its future directions.

**4.1** Pretext tasks of SSL can create implicit supervisory signals from unlabelled datasets to perform unsupervised learning close to, or even equal to, that of human labelling. The pretext tasks we survey are all manually created by experts, and require both domain and machine learning skills, together with a comprehensive set of experiments. We believe there is an opportunity to frame the pretext task creation as an optimisation problem, which is conceptually comparable to the pursuit of the best architecture for a deep learning challenge. Furthermore, learning a reliable representation from medical images will not be optimal by simply adopting pretext tasks that have been developed on natural images. Hence, such methods require to be further modified and improved to suit the nature of medical images and enable extracting robust representation.

**4.2** Similar to pretext tasks, augmentation techniques used in contrastive SSL methods that are designed and optimised for natural images may not be suitable for medical images. As an example, medical images that are already grayscale would not be transformed in a meaningful way by colour jittering or random grey scale, which are common techniques applied to natural images. The effects of various additional augmentations and their combinations should be studied in further research.

**4.3** Sampling strategies are one of the reasons for the success of mutual information-based systems, as noted by Tschannen et al.[44]. Sampling strategies may affect contrastive SSL methods, such as MoCo and SimCLR, that need huge amounts of negative samples. Hence, how to decrease the reliance on sampling strategies is still an appealing and unsolved problem. A suitable negative sample can be built based on the properties of medical images, and from there, more valuable data features can be extracted[225, 226]. There needs to be further investigate how to create negative samples and how to better adapt SSL to downstream tasks to enhance the performance of SSL approaches in the medical imaging domain. Moreover, along with data augmentation, the redesign of the contrastive loss function plays a crucial role in the performance. Some researchers work on designing contrastive loss functions for their particular purposes in medical areas and related to e.g., multimodal learning[138, 139, 227], local representation learning[141], multi-scale learning, and texture[150] or structural[151] representation learning.

# Appendix

**Table 1** Self-supervision: context-instance contrast/predicting spatial relative position

| Pretext task | Author(s) | Dataset(s) used (in pre-training, testing, and downstream tasks) | Application(s) |
|---|---|---|---|
| Relative position | Chen et al., 2019[85] | (1) 2D fetal ultrasound image (2) 3D abdominal CT image (3) Brain T1 MR image (BraTS Challenge) | (1) Fetal standard scan plane classification (2) Abdominal multiorgan localization (3) Brain tumor segmentation |
| | Blendowski et al., 2019[89] | VISCERAL Anatomy3 CT dataset | Multiorgan segmentation (liver, spleen, left kidney, right kidney, left psoas major muscle, and right psoas major muscle) |
| | Jana et al., 2019[90] | (1) MICCAI 2017 LiTS challenge dataset (2) CT images | (1) Fibrosis classification (2) NAS score classification (non-alcoholic fatty liver disease (NAFLD) Activity Scores (NAS)) |
| | Li et al., 2019[91] | Chest CT images | COVID-19 severity level prediction |
| Jigsaw puzzle | Taleb et al., 2020[93] | (1) BraTS dataset (2) Prostate dataset (3) CHAOS multimodal dataset | (1) Survival days prediction, and multimodal brain tumor segmentation (2) Prostate segmentation (3) Liver segmentation |
| | Taleb et al., 2019[94] | BraTS challenge | (1) Brain tumor segmentation (2) Survival prediction regression |
| | Navarro et al., 2021[95] | (1) X-ray images (RSNA) (2) VISCERAL CT dataset (3) Grand Challenges CT dataset | (1) Pneumonia classification (2) Multiorgan segmentation |
| | Manna et al., | MRNet dataset | Three knee conditions classification |

| | | |
|---|---|---|
| | 2021[96] | (abnormality, ACL tear, and meniscus tear) |
| | Li et al., 2020[97] | (1) MoNuSeg dataset<br>(2) ISIC dataset | [Histopathological images]<br>(1) Nuclei segmentation<br>(2) Skin lesion segmentation |
| | Chae et al., 2021[104] | Cervix image dataset | Cervical cancer classification |
| | Santilli et al., 2021[105] | REIMS data | Breast cancer classification |

| Pretext task | Author(s) | Dataset(s) used | Application(s) |
|---|---|---|---|
| Rubik's cube | Zhuang et al., 2019[106] | (1) Brain hemorrhage CT dataset (private dataset)<br>(2) BraTS-2018 | (1) Brain hemorrhage classification<br>(2) Brain tumor segmentation |
| | Zhu et al., 2020[107] | (1) Cerebral hemorrhage dataset<br>(2) BraTS-2018 | (1) Cerebral hemorrhage classification<br>(2) Brain tumor segmentation |
| | Tao et al., 2020[108] | (1) NIH Pancreas CT dataset<br>(2) MRBrainS18 dataset | (1) Pancreas segmentation<br>(2) Brain tissue segmentation |
| | Li et al., 2020[109] | COVID-19 CT dataset | Distinguishing COVID-19 from other two cases: non-pneumonia and community acquired pneumonia (CAP) on chest CT exams |
| Rotation | Tajbakhsh et al., 2019[118] | (1) & (2) LIDC-IDRI chest CTs<br>(3) Diabetic retinopathy (DR) fundus image dataset<br>(4) Private dataset (color, telemedicine) | (1) False-positive reduction (FPR) for nodule detection<br>(2) Lung lobe segmentation<br>(3) DR classification in fundus images<br>(4) Skin segmentation |
| | Li et al., 2021[110] | (1) iChallenge-AMD dataset<br>(2) iChallenge-PM dataset<br>(3) EyePACS dataset/Kaggle DR | Retinal disease classification |
| | Yang et al., 2020[111] | LiTS 2017 MICCAI | Cross-modality liver segmentation |
| | Liu et al., 2019[115] | (1) NLST Dataset<br>(2) LUNA16 Dataset<br>(3) SPIE-AAPM Dataset<br>(4) Lung TIME Dataset<br>(5) HMS Lung Cancer Dataset | Pulmonary nodule classification |
| | Dong et al., 2021[116] | CT images dataset | Focal liver lesions classification |
| | Koohbanani et al., 2020[121] | (1) Camelyon16<br>(2) LNM-OSCC<br>(3) Kather | [histopathology image]<br>Histology image classification |

**Table 2** Self-supervision: context-instance contrast/maximizing mutual information

| Pretext task | Author(s) | Dataset(s) used (in pre-training, testing, and downstream tasks) | Application(s) |
|---|---|---|---|
| Contrastive Predictive Coding (CPC) | Stacke et al., 2020[132] | (1) STL-10<br>(2) CAMELYON17<br>(3) AIDA-LNCO<br>(4) AIDA-SKIN | [Histopathological images]<br>Tumor classification |
| | Taleb et al., 2020[133] | (1) Multimodal Brain Tumor Segmentation (BraTS) 2018<br>(2) Pancreas dataset<br>(3) Diabetic Retinopathy 2019 Kaggle challenge<br>(4) UK Biobank (UKB) | (1) Brain tumor segmentation<br>(2) Pancreas tumor segmentation<br>(3) Diabetic retinopathy detection |

| | Zhu et al., 2021[134] | 3D brain hemorrhage dataset (private dataset) | Brain hemorrhage classification |
|---|---|---|---|
| | Zhu et al., 2020[136] | (1) Brain hemorrhage dataset (private dataset)<br>(2) LUNA16 dataset | (1) Brain hemorrhage classification<br>(2) Lung nodule classification |

**Table 3** Self-supervision: instance-instance contrast/predicting spatial relative position

| Pretext task | Author(s) | Dataset(s) used (in pre-training, testing, and downstream tasks) | Application(s) |
|---|---|---|---|
| PIRL | Kwasigroch et al., 2020[165] | ISIC2017 challenge dataset | Skin lesion classification |
| MoCo | Sowrirajan et al., 2021[166] | (1) CheXpert dataset<br>(2) Shenzhen Hospital X-ray dataset | Chest X-ray interpretation |
| | Sriram et al., 2021[167] | (1) MIMIC-CXR dataset<br>(2) CheXpert<br>(3) NYU COVID dataset | (1) Adverse event prediction from single images (SIP)<br>(2) Adverse event prediction from multiple images (MIP)<br>(3) Oxygen requirements prediction from single images (ORP) |
| | He et al., 2020[168] | They built the COVID19-CT dataset through collecting medical images from COVID-19 relative bioRxiv and medRxiv papers. | Diagnosing COVID-19 from CT scans |
| | Reed et al., 2021[170] | (1) Chexpert<br>(2) Chest-X-ray-kids | (1) Five classification on Chexpert dataset.<br>(2) Singular classification on Chest-X-ray-kids dataset. |
| | Liang et al., 2020[173] | FedCheXpert | Multi-class classification |
| MoCo+ SimCLR | Gazda et al., 2021[187] | (1) CheXpert dataset<br>(2) Cell dataset<br>(3) ChestX-ray14<br>(4) C19-Cohen dataset<br>(5) COVIDGR dataset | (1) & (2) Pneumonia classification<br>(3) & (4) COVID-19 classification |
| SimCLR | Azizi et al., 2021[186] | (1) Dermatology dataset<br>(2) CheXpert dataset | (1) Dermatology skin condition classification<br>(2) Five pathologies chest X-ray classification |
| | Ciga et al., 2020[188] | (1) BACH challenge dataset<br>(2) Patch Camelyon<br>(3) BreakHis<br>(4) NCT-CRC-HE-100K/ Kather<br>(5) PANDA<br>(6) BACH challenge dataset<br>(7) Gleason2019<br>(8) DigestPath2019<br>(9) BreastPathQ dataset | [histopathology images]<br>(1) Breast cancer classification<br>(2) Lymph node classification<br>(3) Breast tumor classification<br>(4) Colorectal cancer classification:<br>(5) Prostate cancer grading<br>(6) Breast cancer segmentation<br>(7) Prostate cancer grading<br>(8) Colon tumor segmentation.<br>(9) Percentage of cancer cellularity of each patch |
| | Li et al., 2021[189] | (1) Camelyon16<br>(2) TCGA lung cancer dataset The Cancer Genome Atlas (TCGA). | [histopathology images]<br>(1) Breast cancer classification and localization<br>(2) lung cancer classification |
| | Inglese et al., 2021[202] | Private dataset for diagnosing NPSLE | NPSLE/non-NPSLE classification |
| | Zheng et al., 2021[203] | (1) LASC<br>(2) LiTS<br>(3) MSD<br>(4) Knee<br>(5) ACDC | Eight medical image segmentation: cardiovascular structures, liver & tumours, spleen, knee bones & cartilages, and prostate. |

| | | (6) M&Ms | |
|---|---|---|---|
| Cluster discrimination | Abbas et al., 2020[204] | (1) Collected from three different dataset (2) COVID-19 dataset-A (3) COVID-19 dataset-B | Detection of COVID-19 cases |
| | Abbet et al., 2020[206] | Kather dataset | [WSIs histopathological images] Colorectal Cancer classification |
| | Chaves et al., 2021[209] | (1) Isic19 (2) Isic20 (3) Derm7pt–derm and derm7pt–clinic (4) Pad-ufes-20 | Skin lesions classification |
| Contrastive loss function-based variation | Holmberg et al., 2019[138] | (1)Kaggle diabetic retinopathy dataset (2)Tissue segmentation infrared (IR) fundus image dataset | (1) Diabetic retinopathy classification (2) OCT retinal thickness measurements |
| | Xie et al., 2020[141] | (1) Collected from public datasets (RibFrac dataset and Medical Segmentation Decathlon (MSD) challenge) (2) Liver dataset (3) Spleen dataset (4) KiTS dataset (5) BCV dataset | Human organs and two tumor, such as liver and kidney segmentation. |
| | Chaitanya et al., 2020[142,143] | (1) ACDC dataset (2) Prostate dataset (3) MMWHS dataset | (1) Cardiac multi-structures segmentation (2) Prostate structures segmentation (3) Heart multi-structures segmentation |
| | Yan et al., 2020[147] | (1) DeepLesion CT datasets (2) NIH-LN (3) ChestCT dataset | (1)3D universal lesion matching on CT. (2)2D landmark detection on hand and pelvic X-rays. (3)3D 19 landmark detection on chest CT |
| | Sahasrabudhe et al., 2020[148] | (1) MoNuSeg (2) TNBC (3) CoNSeP | [Histopathological images] Nuclei Segmentation |
| Specific structure-based | Xie et al., 2020[124] | MoNuSeg 2018 Dataset | [Histopathological images] Nuclei Segmentation |
| | Spitzer et al., 2018[152] | Generated a dataset based on BigBrain | [Histological images] Cytoarchitectonic Segmentation of human brain areas |
| | Li et al., 2021[153] | (1) BraTS (2) NSCLC-radiomics | (1) Brain tumor classification (2) Lung cancer staging |
| | Dhere et al., 2021[155] | KiTS 2019 challenge | Kidney segmentation |
| | Jamaludin et al., 2017[156] | In-house dataset (TwinsUK registry) | Radiological grading classication |
| | Rivail et al., 2019[157] | Longitudinal dataset | Conversion to advanced AMD Classification |
| | Li et al., 2021[158] | Gastric X-ray image dataset | Gastritis detection |
| | Liu et al., 2021[160] | Chest X-ray14 | Thorax disease multi-label classification |
| Temporal Contrast | Gildenblat et al., 2020[213] | Camelyon16 | [Histopathological images] Image retrieval for tumor areas. |
| | Bai et al., 2019[214] | UK Biobank | Cardiac MR image segmentation |
| | Lu et al., 2021[218,219] | HCP dMRI scan dataset | White matter tract segmentation |
| | Nguyen et al., 2020[212-221] | (1)StructSeg dataset (2)RSNA Intracranial hemorrhage is a CT scan dataset | (1)Organ segmentation (2)Intracranial hemorrhage detection |
| | Jiao et al., 2020[222] | Clinical fetal US dataset | Standard plane detection and saliency prediction. |
| | Klinghoffer et al., 2020[223] | (1)SHIELD PVGPe dataset (2)Single neuron Janelia dataset. | Axon Segmentation |
| | Srinidhi et al., | (1)BreastPathQ dataset | (1)Tumor metastasis detection |

| | | |
|---|---|---|
| 2021[224] | (2)Camelyon16 dataset: | (2)Tissue type classification |
| | (3)Kather multiclass dataset | (3)Tumor cellularity quantification |

# Open Access

This article is licensed under a Creative Commons Attribution 4.0 International License, which permits use, sharing, adaptation, distribution, and reproduction in any medium or format, as long as you give appropriate credit to the original author(s) and the source, provide a link to the Creative Commons licence, and indicate if changes were made.

The images or other third-party material in this article are included in the article's Creative Commons licence, unless indicated otherwise in a credit line to the material. If material is not included in the article's Creative Commons licence and your intended use is not permitted by statutory regulation or exceeds the permitted use, you will need to obtain permission directly from the copyright holder.

To view a copy of this licence, visit http://creativecommons.org/licenses/by/4.0/.

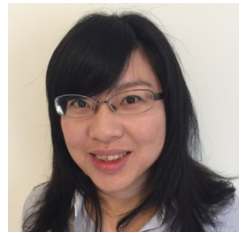

**Wei-Chien Wang** is a PhD candidate at the Biomedical and Multimedia Information Technology (BMIT) Research Group, School of Computer Science, The University of Sydney, Australia.

She received the B.Sc. degree in Mechatronic Engineering from Huafan University, Taipei, Taiwan, China in 2006, and the M.Sc. degree in Manufacturing Information and Systems from the National Cheng Kung University (NCKU), Tainan, Taiwan, China in 2008. She was a full-time research assistant at the National Taiwan Normal University (NTNU), Taipei, Taiwan, China from 2009 to 2010, and a part-time research assistant with Academia Sinica, Taipei, Taiwan, China from 2011.

She worked as a software engineer in Taiwan at Hi-Lo System Research Co., Ltd. from 2010 to 2011, and at the Software Design Center, Foxconn International Holdings, Ltd., Foxconn Technology Group, and the FIH Taiwan Design Center, Hon Hai Precision Industry Co., Ltd. between 2011 and 2012. Since 2013, she has been a research student in Australia, working on various projects in deep learning and computer vision. She is also a visiting researcher and lecturer with the National Penghu University of Science and Technology (NPU), Penghu, Taiwan, China from 2021.

With a history in visual deep learning and Artificial Intelligence of Things (AIoT), now she focuses on self-supervised learning for medical image analysis.

E-mail: wwan7784@uni.sydney.edu.au (Corresponding author)

ORCID iD: 0000-0002-3255-7212

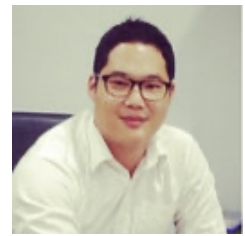

**Euijoon Ahn** is a Lecturer at the College of Science and Engineering, James Cook University, Cairns, Australia. Prior to this, he was a Postdoctoral Research Fellow at the Biomedical and Multimedia Information Technology (BMIT) Research Group, School of Computer Science, The University of Sydney.

He received the B.I.T. degree from the University of Newcastle, Australia in 2009, and the M.I.T. and M.Phil. degrees from the University of Sydney in 2014 and 2016 respectively. He obtained the Ph.D. degree in Computer Science from The University of Sydney in 2020.

Dr. Ahn is a member of SICE, IEE, and IEEE. He produced top-tier publications in computer vision and medical image computing, including papers in *IEEE T-MI, T-BME, JBHI, MedIA, PR, CVPR, AAAI* and *MICCAI*. He is a regular reviewer for *IEEE T-PAMI, T-MI, Nature Communications, CVPR, MICCAI* and *ISBI*.

His research in the development of machine learning and computer vision focuses on unsupervised and self-supervised deep learning models for biomedical image analysis, for improving image segmentation, retrieval, quantification, and classification, without relying on labelled data. He also works in translational health technology research, especially on health data analytics and telehealth.

E-mail: euijoon.ahn@jcu.edu.au

ORCID iD: 0000-0001-7027-067X

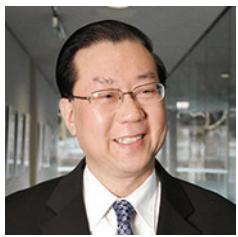

**Dagan Feng** is a Professor Emeritus at the School of Computer Science, The University of Sydney, Australia, and the founding Director of the Biomedical and Multimedia Information Technology (BMIT) Research Group. He is a Fellow of ACS, HKIE, IET, IEEE, and the Australian Academy of Technological Sciences and Engineering (ATSE).

He received the M.Eng. degree in Electrical Engineering and Computer Science (EECS) from Shanghai Jiao Tong University (SJTU), China, in 1982, and the M.Sc. degree in Biocybernetics and the Ph.D. degree in Computer Science from the University of California, Los Angeles (UCLA), in 1985 and 1988, respectively, where he received the Crump Prize for Excellence in Medical Engineering. After briefly working as Assistant Professor at the University of California, Riverside, he joined the University of Sydney at the end of 1988, as Lecturer, progressing onto Professor in the Department of Computer Science and Head of School of Information Technologies.

Prof. Feng has led more than 50 key research projects, published over 700 scholarly research papers, pioneered several new research directions, and made a number of landmark contributions in his field. More importantly, however, is that many of his research results have been translated into solutions to real-life problems and have made tremendous improvements to the quality of life for those concerned.

He has served as Chair of the International Federation of Automatic Control (IFAC) Technical Committee on Biological and Medical Systems, Special Area Editor / Associate Editor / Editorial Board Member for a dozen of core journals in his area, and Scientific Advisor for a number of prestigious organizations. He has been invited to give over 100 keynote presentations in 23 countries and regions, and has organized / chaired over 100 major international conferences and symposia.

He has also been appointed as Honorary Research Consultant, Royal Prince Alfred Hospital in Sydney; Chair Professor of Information Technology, Hong Kong Polytechnic University; Advisory Professor, Shanghai Jiao Tong University; Guest Professor, Northwestern Polytechnic University, Northeastern University, and Tsinghua University.

Prof. Feng's research in the areas of biomedical systems modelling, functional imaging, biomedical information technology, and multimedia computing seeks to address the major challenges in "big data science" and provide innovative solutions for stochastic data acquisition, compression, storage, management, modelling, fusion, visualization, and communication. Currently, Prof. Feng and his research collaborators are working on new ways of improving the early detection of diseases such as cancer and dementia.

E-mail: dagan.feng@sydney.edu.au

ORCID iD: 0000-0002-3381-214X

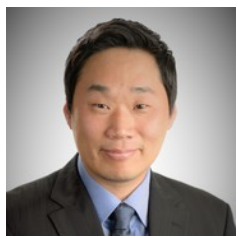

**Jinman Kim** is an Associate Professor of computer science and the founding Director of the Biomedical Data Analysis and Visualisation (BDAV) Lab at the University of Sydney, Australia. He also serves as an Associate Director of the School of Computer Science's Biomedical and Multimedia Information Technology (BMIT) Research Group. He co-leads the "digital health imaging", as part of the Faculty of Engineering's Digital Science Initiative, with the vision and strategy to improve the use and accessibility of medical imaging via AI innovations.

He received the B.S. (Hons.) and Ph.D. degrees in Computer Science from the University of Sydney, Australia, in 2001 and 2006 respectively. Since 2006, he has been a Research Associate with the university's leading teaching hospital, the Royal Prince Alfred Hospital. From 2008 to 2012, he was an ARC Postdoctoral Research Fellow, with one year leave from 2009 to 2010 to join the MIRALab Research Group, Geneva, Switzerland, as a Marie Curie Senior Research Fellow. Since 2013, he has been with the University of Sydney School of Computer Science, where he was a Senior Lecturer, and was promoted to Associate Professor in 2016.

He continuously publishes in top venues in his field and received multiple competitive grants and scientific recognitions. He is actively involved in his research communities where he is the Vice President of the Computer Graphics Society (CGS), A/Editor of *Computer Methods and Program in Biomedicine (CMPB)*, A/Editor of *The Visual Computer (TVCJ)*, and Reviewer for all major journals and conferences in his field.

He has actively focused on research translation where he has worked closely with clinical partners to take his research into clinical practice. He is the Research Director of the Nepean Telehealth and Technology Centre (NTTC) at Nepean Hospital, NSW Health, responsible for translational telehealth and digital hospital research. Some of his research has been developed into clinical software that is being used or trialled at multiple hospitals. His work on telehealth has been recognised with multiple awards, including the 2016 Health Secretary Innovation Award from the NSW Ministry of Health.

Prof. Kim's research is in the intersection of machine learning, and biomedical image analysis and visualization, especially on multi-modal data processing, image-omics, and image data correlation to other health data.

E-mail: jinman.kim@sydney.edu.au

ORCID iD: 0000-0001-5960-1060